\documentclass[letterpaper, 10 pt, conference]{ieeeconf}  % Comment this line out if you need a4paper

\IEEEoverridecommandlockouts                              % This command is only needed if 
\usepackage{cite}
\usepackage{amsmath,amssymb,amsfonts}
\usepackage{algorithmic}
\usepackage{graphicx}
\usepackage{textcomp}
\usepackage{xcolor}
\usepackage{booktabs}
\usepackage{multirow}

\title{\LARGE \bf
PASS-VLN: Passability-Aware Local-Command \\Vision-and-Language Navigation
}

\author{
Cong Zou$^{1}$,
Xiangsheng Shi$^{1}$,
Xuanhui Wang$^{1}$,
Fubin Wang$^{1}$,
Guangjun Ge$^{1}$,
Chao Wang$^{1,*}$,
Yu Wang$^{1,2}$%
\thanks{$^{1}$Qiyuan Lab, Beijing, China.}%
\thanks{$^{2}$Department of Electronic Engineering,
Tsinghua University, Beijing, China.}%
\thanks{$^{*}$Corresponding author: Chao Wang
(\texttt{wangchao1@qiyuanlab.com}).}%
}

\begin{document}

\maketitle
\thispagestyle{empty}
\pagestyle{empty}

\begin{abstract}

Reliable vision-and-language navigation (VLN) requires not only understanding instructions but also determining whether and how a robot can physically execute them. However, existing VLN benchmarks typically assume executable instructions and focus on long-horizon tasks. This leaves body-dependent local execution, including passability and collision-free execution, largely unexamined. Evaluating these factors within long-horizon tasks would entangle local physical execution with task planning and long-term spatial memory, making navigation failures difficult to attribute. We therefore focus on grounding local navigation commands in executable motion under environmental and robot body constraints. We present PASS-VLN, a passability-aware local-command VLN benchmark and model. The benchmark holds the scene, instruction, start pose, and reference route fixed while varying robot body width, requiring target completion when the instructed passage is feasible and stopping before blockage otherwise. Our model encodes body width through obstacle inflation in a depth-derived local bird’s-eye view, fuses geometry with visual and linguistic context, and predicts discrete latent action codes for continuous closed-loop execution. Experiments show that PASS-VLN improves collision-free success by approximately $12$ percentage points over a strong VLN model fine-tuned on our benchmark, while adapting reach-versus-stop behavior to body constraints and generalizing to unseen body widths. When integrated with a high-level planner, PASS-VLN achieves higher success with fewer local commands, while real-world deployment further demonstrates its effectiveness as a low-level executor. The benchmark and model will be released.

\end{abstract}

\section{Introduction}
Vision-and-Language Navigation (VLN) requires an embodied agent to follow natural-language instructions and navigate to a target in visual environments. R2R established the standard indoor VLN setting~\cite{anderson2018vision}, followed by benchmarks for more complex multilingual instructions~\cite{ku2020room}, continuous control~\cite{krantz2020beyond}, remote object grounding~\cite{qi2020reverie}, persistent navigation~\cite{krantz2023iterative}, and physics-aware unified evaluation~\cite{lin2025vlnverse}.

Most existing VLN benchmarks provide a complete route instruction per episode and assume that the instructed motion is physically executable. Directly incorporating body-dependent physical execution, including passability and collision-free execution, would couple local execution with planning, progress tracking, spatial memory, and route-level reasoning, making navigation failures difficult to attribute. Many recent VLN methods emphasize stronger reasoning, memory, and planning~\cite{rajvanshi2024saynav,zhou2024navgpt,zheng2024navillm,qiao2025opennav,lin2026evolvenav, xin2026agentvln,wang2026reflectvln,guo2026awarevln}, whose gains do not necessarily translate into improved local physical execution. Conventional navigation methods based on local mapping and motion planning account for traversability and collision constraints, but typically take geometric goals or reference paths as input without a natural-language interface for interpreting spatially grounded commands~\cite{hart1968formal,fox1997dynamic,rosmann2013efficient}.

These considerations motivate focusing on local physical execution as a capability that complements high-level reasoning and memory. A human or a high-level agent, including one based on a VLM, can issue local commands, while a low-level executor grounds each command in current observations and converts it into physical motion. The local command serves as their semantic interface, specifying an immediate local objective and the spatial cues needed to interpret it. For example, ``move forward and stop between the sofa and the table'' specifies landmarks, spatial relations, and a stopping condition while leaving broader task planning to the command issuer.

Such a local executor must reason about not only \emph{where} to go, but also \emph{whether and how} the motion can be physically executed. Most existing VLN research emphasizes semantic spatial reasoning, such as left/right relations, object ordering, relative positions, and path structure~\cite{wang2023gridmm,huo2023geovln,dai2024optimal,zhang2025mapnav,
zhu2025mtu3d,zeng2025janusvln}, without explicitly considering physical spatial feasibility under body constraints. In practice, a correctly grounded passage may still be too narrow for the robot, and traversable passages may require body-dependent clearance and alignment. A local executor should therefore consider the instructed motion and its physical feasibility under the current embodiment.

To address these requirements, we introduce \textbf{PASS-VLN}, a  \textbf{PASS}ability-aware local-command \textbf{VLN} framework comprising a controlled benchmark and a body-width-conditioned model for planar physical execution. The benchmark holds the scene, instruction, start pose, and reference route fixed while varying robot body width. For paired configurations where passage feasibility changes under the same command, the narrower robot is expected to reach the target, whereas the wider robot is expected to stop before the blockage, directly testing whether execution adapts to the interaction between instructed motion, environmental geometry, and robot embodiment. The benchmark covers commands based on local observations and commonsense spatial knowledge for visible-target navigation or bounded exploration, as well as commands requiring broader scene or route knowledge.

To connect language understanding with body-dependent geometry and motion execution, PASS-VLN projects depth into a local bird’s-eye view (BEV) and incorporates body width through obstacle inflation, making the geometric representation reflect robot-specific clearance. Fusing this representation with language, current vision, and visual history allows the policy to interpret the instructed motion in relation to its physical feasibility. To translate this understanding into continuous execution, the model predicts discrete latent action codes learned from short trajectory windows, which are decoded into planar relative motions for closed-loop control. The main contributions are summarized as follows:

\begin{itemize}
    \item We introduce a controlled VLN benchmark that isolates local execution from long-horizon navigation and explicitly evaluates body-width-dependent physical feasibility. Its paired counterfactual design tests whether an agent changes its reach-versus-stop behavior under different body constraints while preserving the command’s semantic intent.

    \item We develop a body-width-conditioned VLN model for local physical execution. Body-conditioned geometry allows the same policy to adapt passability decisions and motion execution across seen and unseen widths, while discrete latent actions preserve VLM-style token prediction and enable continuous closed-loop motion.

    \item Experiments demonstrate generalization to unseen body widths and improved collision-free success over the evaluated binary passability-gating baseline. In a shared reference-route-assisted hierarchical pipeline, PASS-VLN achieves higher collision-free navigation success with fewer local commands, while real-world deployment further validates its practical utility.
\end{itemize}

\section{Related Work}
\label{sec:related_work}

\begin{figure*}[t]
    \centering
    \includegraphics[width=1.0\textwidth]{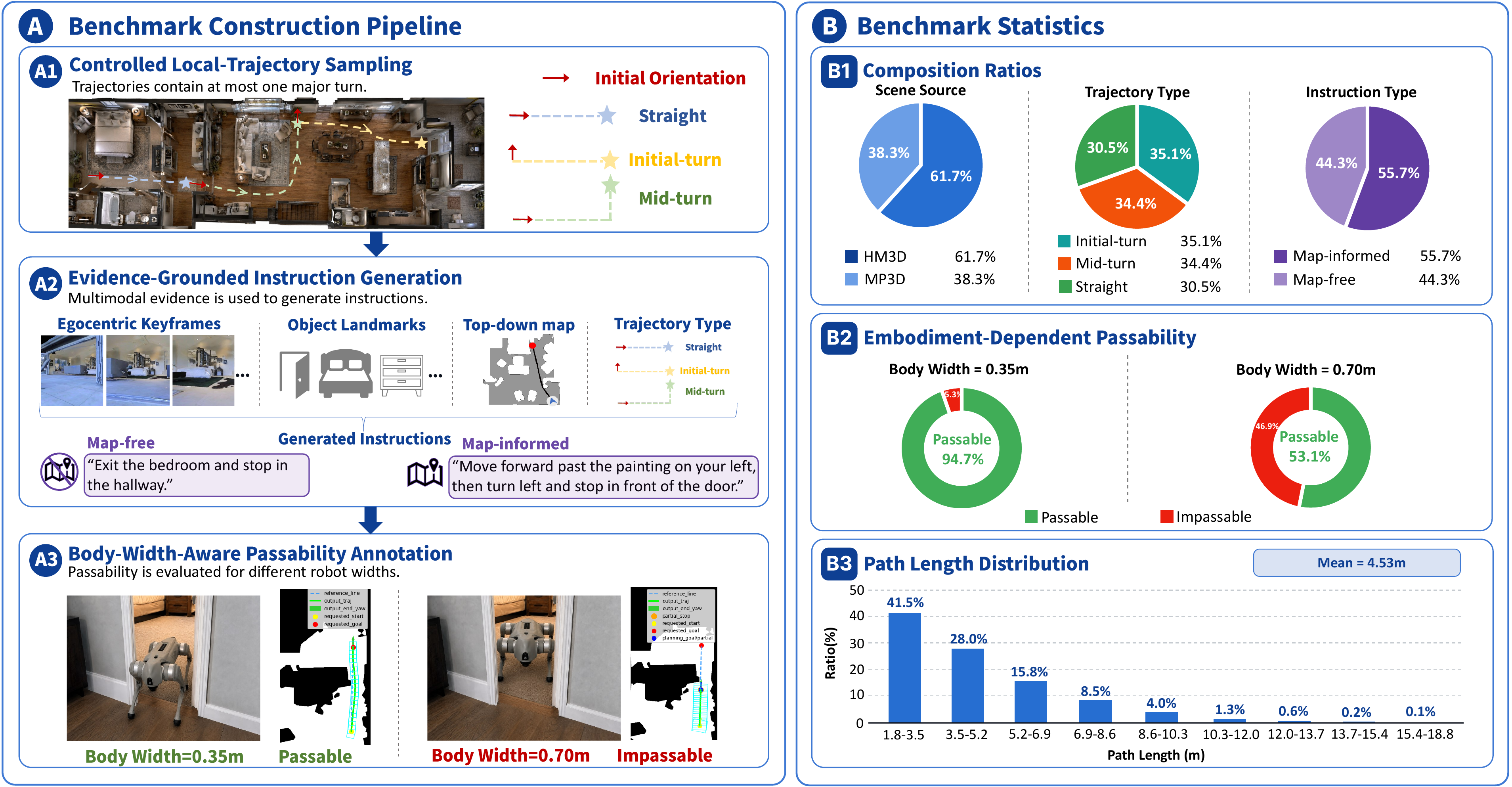}
    \caption{Construction pipeline and statistics of the PASS-VLN benchmark. Controlled local trajectories are paired with multimodal evidence to generate commands with different spatial-context requirements, followed by body-width-aware passability annotation.}
    \label{fig:data_pipeline}
\end{figure*}

\subsection{Vision-and-Language Navigation Benchmarks}
\label{subsec:vln_benchmarks}

R2R introduced instruction following in photorealistic indoor environments~\cite{anderson2018vision}, while RxR increased linguistic diversity through multilingual instructions and dense spatiotemporal grounding~\cite{ku2020room}. VLN-CE moved navigation from discrete graphs to continuous environments with low-level actions~\cite{krantz2020beyond}. REVERIE combined navigation with remote object grounding~\cite{qi2020reverie}, and Touchdown studied navigation and spatial reasoning in outdoor street-view environments~\cite{chen2019touchdown}. AerialVLN extended the task to UAV navigation~\cite{liu2023aerialvln}, and VLNVerse introduced a scalable physics-aware framework covering multiple VLN tasks and embodied agents~\cite{lin2025vlnverse}. PASS-VLN complements these benchmarks by isolating local-command spatial grounding and explicitly evaluating whether the local motion specified by an instruction remains physically executable under different body widths, including unseen widths.

\subsection{Large-Model-Based VLN and Physical Execution}
\label{subsec:vln_models}

Recent VLN methods increasingly use large language and vision-language models for navigation reasoning and action generation. StreamVLN introduces slow--fast context modeling for efficient streaming navigation~\cite{wei2025streamvln}, while JanusVLN uses separate visual-semantic and spatial-geometric memories to improve spatial reasoning~\cite{zeng2025janusvln}. CorrectNav introduces a self-correction flywheel that learns from navigation-error trajectories \cite{yu2026correctnav}, while AwareVLN performs structural self-aware reasoning over agent state and task progress \cite{guo2026awarevln}. Recent work also considers embodiment and physical execution: NaVILA combines VLA planning with legged locomotion skills~\cite{cheng2025navila}, VAMOS introduces embodiment-specific affordance evaluation~\cite{guamancastro2025vamos}, DualVLN couples high-level visual-language reasoning with a continuous local diffusion policy~\cite{wei2025dualsystem,cai2026navdp}, and CapNav studies heterogeneous mobility and physical constraints~\cite{su2026capnav}. These methods mainly study route-level navigation, general embodied reasoning, or adaptation across different robot platforms and mobility types. In contrast, PASS-VLN studies reliable local physical execution for hierarchical navigation, using local commands as the interface between high-level reasoning and low-level motion execution, while explicitly considering body-width-dependent physical passability.

\section{PASS-VLN Benchmark}
\label{sec:benchmark}

\subsection{Controlled Local-Trajectory Sampling}
\label{subsec:trajectory_sampling}

We define a local command as a bounded local navigation objective issued by a high-level planner according to the current robot state. It requires the executor to jointly ground landmark semantics, spatial relations, and select environment-conditioned actions, e.g., ``turn left after passing the table'' rather than merely execute isolated directional commands. To construct a controlled evaluation setting, we restrict each trajectory to at most one major heading change, defined as a heading variation greater than $30^\circ$. As shown in Fig.~\ref{fig:data_pipeline}.A1, we sample navigable start--goal pairs in Habitat and categorize each path as straight, initial-turn, or mid-turn, corresponding to forward motion, an early turn followed by forward motion, or a turn near an intermediate waypoint. These patterns cover common local navigation structures while limiting repeated turns, long-term progress tracking, and accumulated execution errors, making them suitable for local-command execution in hierarchical navigation.

\subsection{Evidence-Grounded Instruction Generation}
\label{subsec:instruction_generation}

For each candidate trajectory, we construct a multimodal evidence pool of chronological egocentric keyframes, a top-down map, semantic object identities, and a trajectory type, as shown in Fig.~\ref{fig:data_pipeline}.A2. Since a high-level command issuer may have access to different levels of spatial context, we consider two complementary command types. This distinction concerns the context available to the command issuer; the executor receives the instruction and onboard observations in both settings:

\begin{itemize}
    \item \textbf{Map-free commands} can be formulated without map- or route-level information. They rely on current observations and commonsense spatial knowledge, covering both directly visible targets and locally discoverable goals, such as locating a nearby doorway to leave a room.

    \item \textbf{Map-informed commands} require map- or route-level spatial context to specify the intended local motion more explicitly, including turns, landmarks, and stopping locations that may lie beyond the current view and commonsense knowledge.
\end{itemize}

These two settings cover local goal discovery and precise route execution. A VLM generates candidate instructions from the multimodal evidence, followed by geometric and VLM-based verification for grounding, executability, brevity, and unambiguous motion specification.

\subsection{Body-Width-Aware Passability Annotation}
\label{subsec:passability_annotation}

A local command feasible for a narrow agent may be infeasible for a wider robot. PASS-VLN therefore evaluates the instructed local motion under multiple body-width configurations. Passability is defined with respect to the issued command. For map-informed commands, the instructed passage is explicitly specified. For map-free commands, we discard impassable episodes if width-aware replanning finds an alternative route less than $5\,\mathrm{m}$ longer than the original path, indicating a comparable local detour.
 
For each body width, we first perform width-aware path search along the intended route on the top-down traversability map using width-dependent obstacle inflation as illustrated in Fig.~\ref{fig:data_pipeline}.A3, and then validate the resulting trajectory on the original-resolution map using the exact rotated rectangular footprint at sampled poses and intermediate transitions. If the goal is unreachable, the longest collision-free prefix is retained, and its terminal pose defines the pre-blockage stopping reference. Consequently, the success condition depends on body width: the agent should reach the target when the path is traversable and stop before the blockage otherwise.

\subsection{Benchmark Configuration and Metrics}
\label{subsec:benchmark_configuration}

Instruction generation uses Gemma 4 31B~\cite{gemmateam2026gemma4}. The robot footprint is modeled as a rectangle with a fixed longitudinal length of $0.60\,\mathrm{m}$ and a variable body width. We use $0.35\,\mathrm{m}$ and $0.70\,\mathrm{m}$ as training widths, while $0.20\,\mathrm{m}$ and $0.50\,\mathrm{m}$ are held out for evaluating generalization to unseen body
widths.

The PASS-VLN training set contains approximately 192K episodes collected from HM3D and MP3D training scenes. The distributions of scene sources, trajectory types, instruction types, path lengths, and passability under different body widths are summarized in Fig.~\ref{fig:data_pipeline}.B. Evaluation is conducted on MP3D val\_unseen scenes disjoint from the training environments, with $2{,}971$ map-free and $3{,}470$ map-informed episodes. We further conduct human validation on 500 randomly sampled instructions, including $250$ map-free and $250$ map-informed commands, of which $95.6\%$ were judged correct and unambiguous.

\begin{figure*}[htbp]
\centering
\includegraphics[width=0.95\textwidth]{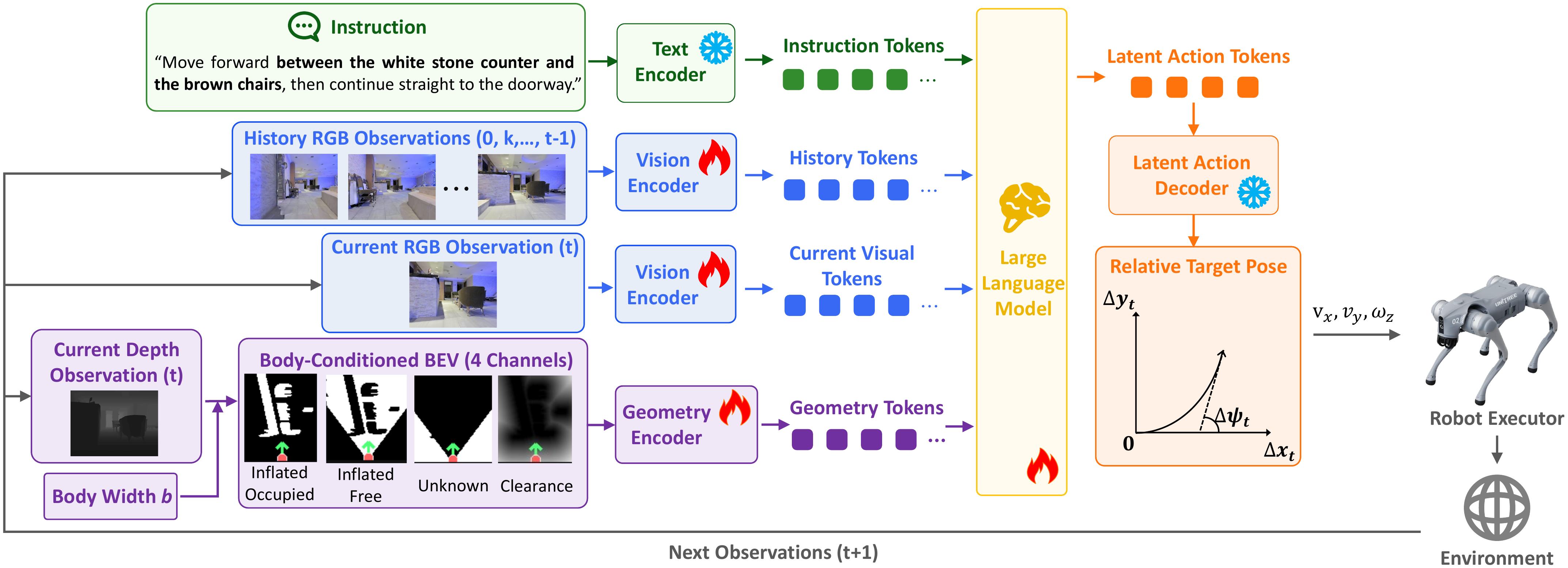}
\caption{Overview of the PASS-VLN architecture. Language, visual history, and body-width-conditioned geometric features are fused to predict latent actions, which are decoded into planar target poses for closed-loop velocity control.}
\label{fig:model_arch}
\end{figure*}

Since PASS-VLN focuses on local physical execution rather than long-horizon tasks, we use a stricter $1.5\,\mathrm{m}$ success radius than the $3.0\,\mathrm{m}$ criterion commonly adopted in R2R-CE. A passable episode is successful if the agent stops within $1.5\,\mathrm{m}$ of the trajectory endpoint, whereas an impassable episode is successful if it stops within the $1.5\,\mathrm{m}$ region immediately before the first blockage. Each episode is limited to $100$ execution steps. 

We report Success Rate (SR), Success weighted by Path Length (SPL), Oracle Success Rate (OSR), and Navigation Error (NE)~\cite{anderson2018vision}. For passable episodes, the target is the trajectory endpoint, while for impassable episodes it is the valid pre-blockage region along the reference path. We additionally report \textbf{Goal Success Rate (GSR)}, the success rate on passable episodes; \textbf{Blockage Success Rate (BSR)}, the success rate on impassable episodes; \textbf{Paired Success Rate (PSR)}, the fraction of paired routes where the narrow robot successfully reaches the target and the wide robot correctly stops before blockage. Since task success alone does not capture the quality of physical execution, we further report \textbf{Collision-Free Rate (CFR)}, the fraction of collision-free episodes, and \textbf{Collision-Free Success Rate (CFSR)}, the fraction of episodes that are both successful and collision-free.

\section{PASS-VLN Model}
\label{sec:model}

\subsection{Model Overview}
\label{subsec:model_overview}

PASS-VLN is a streaming VLN model that jointly processes language instructions, historical and current visual observations, and current body-width-conditioned geometry. As illustrated in Fig.~\ref{fig:model_arch}, the model contains a visual encoder, a body-width-conditioned geometry encoder, a causal multimodal backbone, and a latent action decoder. At each navigation step, PASS-VLN predicts four discrete latent-action tokens. These tokens are decoded into a relative target transformation, which is executed by a low-level controller. The model then acquires new observations and repeats the prediction process in a closed loop.

\subsection{Body-Width-Conditioned Geometry Fusion}
\label{subsec:geometry_fusion}

RGB observations provide semantic and geometric cues, but do not explicitly encode passability with respect to a particular robot body width. PASS-VLN therefore projects the current depth observation onto a robot-centric bird's-eye-view grid. Valid depth points within the configured spatial and height ranges are used to construct a binary occupancy map $\mathbf{O}_t$. To incorporate the robot body width, occupied cells are expanded along the lateral axis using

\begin{equation}
    r_b
    =
    \left\lceil
        \frac{b/2+m}{s_{\mathrm{lat}}}
    \right\rceil,
    \qquad
    \mathbf{O}_{t}^{(b)}
    =
    \operatorname{MaxPool}_{1\times(2r_b+1)}
    \left(
        \mathbf{O}_t
    \right),
\end{equation}
where $b$ is the body width, $m$ is the safety margin, and $s_{\mathrm{lat}}$ is the lateral size of a BEV cell. Since the longitudinal footprint is fixed, only the lateral dimension is explicitly conditioned; longitudinal effects are handled by the fixed dynamics and trajectory supervision as explained in Section~\ref{subsec:passability_annotation}.

The resulting body-conditioned BEV contains four channels: inflated occupancy, body-conditioned free space, unknown space, and normalized obstacle clearance. A lightweight convolutional encoder followed by adaptive spatial pooling converts the BEV into geometry tokens $G_t$, while preserving coarse near--far and left--right structure. Because body width is not provided as a separate scalar but instead reflected through a BEV that can be recomputed for arbitrary widths, the learned policy can operate on unseen body widths without retraining.

Historical RGB observations are encoded as visual memory tokens $\mathbf{M}_t$, while the current RGB observation is encoded as
$\mathbf{V}_{t}^{\mathrm{cur}}$. The geometry tokens are concatenated with the current visual tokens and jointly processed with the instruction and visual history:

\begin{equation}
    \mathbf{H}_t
    =
    F_{\theta}
    \left(
        \mathbf{L},
        \mathbf{M}_t,
        \left[
            \mathbf{G}_t;
            \mathbf{V}_{t}^{\mathrm{cur}}
        \right]
    \right),
\end{equation}
where $F_{\theta}$ denotes the causal multimodal backbone, $\mathbf{L}$ denotes the instruction tokens and $\mathbf{H}_t$ is the multimodal representation used for action prediction.

\begin{table*}[t]
    \centering
    \caption{Local-command navigation performance with a fixed $0.35\,\mathrm{m}$ body width. ZS evaluates released VLN checkpoints without PASS-VLN-specific training. StreamVLN-FT serves as the controlled task-adapted baseline.}
    \label{tab:local_command_transfer}

    \footnotesize
    \setlength{\tabcolsep}{2.8pt}
    \renewcommand{\arraystretch}{0.86}

    \begin{tabular}{@{}lccccccc|cccccc@{}}
        \toprule
        &
        &
        \multicolumn{6}{c|}{\textbf{Map-informed}}
        &
        \multicolumn{6}{c}{\textbf{Map-free}} \\
        \cmidrule(lr){3-8}
        \cmidrule(l){9-14}
        \textbf{Executor} & \textbf{Adaptation} & \textbf{SR}$\uparrow$ & \textbf{OSR}$\uparrow$ & \textbf{SPL}$\uparrow$ & \textbf{NE}$\downarrow$ & \textbf{CFR}$\uparrow$ & \textbf{CFSR}$\uparrow$ & \textbf{SR}$\uparrow$ & \textbf{OSR}$\uparrow$ & \textbf{SPL}$\uparrow$ & \textbf{NE}$\downarrow$ & \textbf{CFR}$\uparrow$ & \textbf{CFSR}$\uparrow$ \\
        \midrule
        StreamVLN & ZS & $29.1$ & $79.8$ & $23.8$ & $3.62$ & $43.2$ & $15.6$ & $31.3$ & $79.8$ & $27.1$ & $3.54$ & $47.2$ & $17.6$\\
        DualVLN & ZS & $29.3$ & $68.7$ & $23.9$ & $3.74$ & $66.2$ & $23.6$ & $35.6$ & $73.3$ & $30.2$ & $3.46$ & $69.8$ & $27.5$\\
        JanusVLN & ZS & $41.3$ & $88.1$ & $35.4$ & $3.33$ & $47.4$ & $24.6$ & $44.2$ & $86.1$ & $39.5$ & $3.48$ & $56.4$ & $30.7$\\
        \midrule
        StreamVLN & FT & $77.7$ & $89.9$ & $72.7$ & $1.14$ & $81.1$ & $64.4$ & $82.5$ & $89.2$ & $81.9$ & $1.17$ & $76.1$ & $63.8$\\
        PASS-VLN & FT & $\mathbf{83.4}$ & $\mathbf{93.7}$ & $\mathbf{75.2}$ & $\mathbf{1.02}$ & $\mathbf{90.6}$ & $\mathbf{76.1}$
        & $\mathbf{85.2}$ & $\mathbf{91.9}$ & $\mathbf{83.0}$ & $\mathbf{1.06}$ & $\mathbf{89.1}$ & $\mathbf{76.0}$ \\
        \bottomrule
    \end{tabular}
\end{table*}

\begin{table*}[t]
    \centering
    \caption{Body-width generalization on seen (S) and unseen (U) widths. GT Passability-Gated VLN uses a narrow ($0.10\,\mathrm{m}$) navigator with simulator-provided passability.}
    \label{tab:width_generalization}

    \footnotesize
    \setlength{\tabcolsep}{1.8pt}
    \renewcommand{\arraystretch}{0.84}

    \begin{tabular}{@{}lcccccccc|ccccccc@{}}
        \toprule
        &&
        \multicolumn{7}{c|}{\textbf{Map-informed}}
        & \multicolumn{7}{c}{\textbf{Map-free}} \\
        \cmidrule(lr){3-9}
        \cmidrule(l){10-16}
        \textbf{Width}
        & \textbf{Method}
        & \textbf{Imp.\%}
        & \textbf{SR}$\uparrow$
        & \textbf{GSR}$\uparrow$
        & \textbf{BSR}$\uparrow$
        & \textbf{CFR}$\uparrow$
        & \textbf{CFSR}$\uparrow$
        & \textbf{NE}$\downarrow$
        & \textbf{Imp.\%}
        & \textbf{SR}$\uparrow$
        & \textbf{GSR}$\uparrow$
        & \textbf{BSR}$\uparrow$
        & \textbf{CFR}$\uparrow$ 
        & \textbf{CFSR}$\uparrow$ 
        & \textbf{NE}$\downarrow$ \\
        \midrule
        \multirow[c]{2}{*}{$0.20$ (U)} & PASS-VLN
        & $1.9$ & $83.3$ & $83.3$ & $83.1$ & $\mathbf{97.2}$ & $\mathbf{81.1}$ & $\mathbf{1.01}$
        & $1.8$ & $\mathbf{85.2}$ & $\mathbf{85.2}$ & $\mathbf{84.9}$ & $\mathbf{97.3}$ & $\mathbf{83.6}$ & $\mathbf{1.05}$ \\
        & GT Passability-Gated VLN
        & $1.9$ & $\mathbf{83.6}$ & $\mathbf{83.6}$ & $\mathbf{87.7}$ & $94.4$ & $80.0$ & $1.03$
        & $1.8$ & $84.4$ & $84.4$ & $\mathbf{84.9}$ & $96.7$ & $82.7$ & $1.10$ \\
        \midrule
        \multirow[c]{2}{*}{$0.35$ (S)} & PASS-VLN
        & $3.4$ & $\mathbf{83.4}$ & $\mathbf{83.4}$ & $82.4$ & $\mathbf{90.6}$ & $\mathbf{76.1}$ & $\mathbf{1.02}$
        & $2.2$ & $\mathbf{85.2}$ & $\mathbf{85.4}$ & $78.5$ & $\mathbf{89.1}$ & $\mathbf{76.0}$ & $\mathbf{1.06}$ \\
        & GT Passability-Gated VLN
        & $3.4$ & $82.7$ & $82.6$ & $\mathbf{83.2}$ & $83.7$ & $71.5$ & $1.07$
        & $2.2$ & $84.7$ & $84.8$ & $\mathbf{80.0}$ & $85.4$ & $73.8$ & $1.11$ \\
        \midrule
        \multirow[c]{2}{*}{$0.50$ (U)} & PASS-VLN
        & $10.6$ & $81.7$ & $81.7$ & $81.5$ & $\mathbf{86.4}$ & $\mathbf{68.7}$ & $\mathbf{1.03}$ 
        & $17.7$ & $83.4$ & $83.8$ & $81.6$ & $\mathbf{84.2}$ & $\mathbf{69.9}$ & $\mathbf{1.06}$ \\
        & GT Passability-Gated VLN
        & $10.6$ & $\mathbf{83.2}$ & $\mathbf{82.5}$ & $\mathbf{89.7}$ & $71.8$ & $63.0$ & $1.04$
        & $17.7$ & $\mathbf{84.4}$ & $\mathbf{84.5}$ & $\mathbf{84.0}$ & $71.6$ & $62.3$ & $1.12$\\
        \midrule
        \multirow[c]{2}{*}{$0.70$ (S)} & PASS-VLN
        & $30.4$ & $81.4$ & $79.5$ & $85.8$ & $\mathbf{86.0}$ & $\mathbf{69.2}$ & $1.02$ 
        & $51.6$ & $84.8$ & $81.5$ & $87.8$ & $\mathbf{87.1}$ & $\mathbf{74.7}$ & $0.89$ \\
        & GT Passability-Gated VLN
        & $30.4$ & $\mathbf{84.4}$ & $\mathbf{83.2}$ & $\mathbf{87.3}$ & $47.3$ & $42.5$ & $\mathbf{0.92}$ 
        & $51.6$ & $\mathbf{87.1}$ & $\mathbf{84.0}$ & $\mathbf{90.0}$ & $35.6$ & $31.8$  & $\mathbf{0.85}$\\
        \bottomrule
    \end{tabular}
\end{table*}

\subsection{Latent Action Learning and Execution}
\label{subsec:latent_action}

Instead of predicting fixed primitive actions, PASS-VLN represents a short window of planar motion using discrete latent-action tokens. Let $T_i\in SE(2)$ denote the robot pose, with $\Delta T_i=T_i^{-1}T_{i+1}$ parameterized by $(\Delta x_i,\Delta y_i,\Delta\psi_i)$. Trajectories are collected under maximum linear and angular velocities of $0.5\,\mathrm{m/s}$ and $30^\circ/\mathrm{s}$, respectively, and each $1\,\mathrm{s}$ segment is encoded into two translation and two rotation codes
using separately trained vector-quantized temporal encoders: $\mathbf{c}_t=\left[c_{t,1}^{T}, c_{t,2}^{T}, c_{t,1}^{R}, c_{t,2}^{R}\right]$. The corresponding codebook embeddings are decoded into a planar relative target pose $(\Delta\hat{x}_t,\Delta\hat{y}_t,\Delta\hat{\psi}_t)$. The latent action model is trained with translation reconstruction, rotation geodesic, and vector-quantization losses:

\begin{equation}
    \mathcal{L}_{\mathrm{LAM}}
    =
    \lambda_T\mathcal{L}_T
    +
    \lambda_R\mathcal{L}_R
    +
    \mathcal{L}_{\mathrm{VQ}}.
\end{equation}
After latent-action learning, the code indices are added to the output vocabulary as special tokens, while the decoder and codebooks are frozen during VLN training. Following StreamVLN~\cite{wei2025streamvln}, PASS-VLN retains a sliding-window KV cache for recent observation--action dialogues and compresses older visual context into slow memory tokens. The four latent-action codes are autoregressively predicted as

\begin{equation}
\mathcal{L}_{\mathrm{VLN}}
=
-\sum_{j=1}^{4}
\log p_{\theta}
\left(
c_{t,j}^{*}
\mid
c_{t,<j}^{*},
\mathbf{H}_t,
\mathbf{K}_{t}
\right),
\end{equation}
where $c_{t,j}^{*}$ is the ground-truth latent-action code and $\mathbf{K}_{t}$ denotes the sliding-window KV cache of recent dialogues.

At inference time, the decoded target is converted into body-frame linear velocities $(v_x,v_y)$ and yaw angular velocity $\omega_z$ for closed-loop execution. We interpret the decoded transformation as \textsc{STOP} when
$\sqrt{\Delta \hat{x}_t^2+\Delta \hat{y}_t^2}<0.05~\mathrm{m}
\quad \text{and} \quad
|\Delta \hat{\psi}_t|<3^\circ$,
eliminating the need for a separate termination head. This latent representation retains the token-prediction paradigm of VLMs while enabling smooth displacement, heading adjustment, passage alignment, and closed-loop obstacle avoidance.

\section{Experiments}
\label{sec:experiments}

\subsection{Implementation Details}
\label{subsec:experimental_setup}

PASS-VLN follows the streaming navigation architecture of StreamVLN~\cite{wei2025streamvln} because its efficient streaming design is well suited for a low-level VLN executor, while introducing body-width-aware geometric reasoning and latent-action prediction. The multimodal backbone is initialized from LLaVA-Video 7B~\cite{zhang2025llavavideo}. We first fine-tune the model on the R2R-CE and RxR-CE training sets to acquire general VLN capability, and then further fine-tune it on the PASS-VLN training set with the body-width settings in Sec.~\ref{subsec:benchmark_configuration}. Training is performed on $8$ NVIDIA A800 GPUs for one epoch with a batch size of $128$, an initial learning rate of $2\times10^{-5}$, cosine decay with $7.5\%$ warmup, and BF16 precision. Each latent action uses two translation and two rotation codes, with a codebook size of $8$. The local BEV is represented as a $96\times96$ grid covering $4.5\,\mathrm{m}$ forward and $\pm2.5\,\mathrm{m}$ laterally, and is encoded into $16$ geometry tokens. The evaluation protocol and metrics follow Sec.~\ref{subsec:benchmark_configuration}. Unless otherwise stated, all controlled variants use the same backbone, optimization protocol, and low-level controller.

\subsection{Local-Command Execution}
\label{subsec:local_command_transfer}

We first compare PASS-VLN with StreamVLN, DualVLN, and JanusVLN on local-command navigation using their released checkpoints under a fixed body width of $0.35\,\mathrm{m}$. We additionally fine-tune StreamVLN on the PASS-VLN training set as a controlled adaptation baseline, since PASS-VLN shares the same base streaming architecture, enabling a controlled comparison of task adaptation and model design.

As shown in Table~\ref{tab:local_command_transfer}, PASS-VLN achieves the best overall performance among the evaluated methods. The large OSR--SR gaps of the directly transferred VLN models also show that strong route-level navigation performance does not necessarily lead to precise local completion and stopping. Fine-tuning StreamVLN substantially improves its performance, confirming the importance of local-command supervision, while PASS-VLN further improves physical execution through body-width-aware geometry and latent-action prediction.

\subsection{Body-Width-Aware Navigation}
\label{subsec:body_width_evaluation}

We next evaluate whether PASS-VLN adapts its navigation behavior to robot body width, using seen widths ($0.35,0.70\,\mathrm{m}$) and unseen widths ($0.20,0.50\,\mathrm{m}$). The $0.50\,\mathrm{m}$ setting tests interpolation between training widths, while $0.20\,\mathrm{m}$ evaluates extrapolation beyond the training range. Passability labels are recomputed for each body width using the annotation method in Section~\ref{subsec:passability_annotation}.

\begin{table}[t]
    \centering
    \caption{Paired body-width evaluation on routes that are passable for the narrower body and impassable for the wider body.}
    \label{tab:paired_width}
    \footnotesize
    \setlength{\tabcolsep}{2.3pt}
    \renewcommand{\arraystretch}{0.87}

    \begin{tabular}{@{}lcccc@{}}
        \toprule
        \textbf{Width Pair} & \textbf{Pair Num.} & \textbf{Narrow GSR}$\uparrow$ & \textbf{Wide BSR}$\uparrow$ & \textbf{PSR}$\uparrow$ \\
        \midrule
        $0.20\ \& \ 0.35\,\mathrm{m}$ & $87$ & $65.5$ & $80.5$ & $64.4$ \\
        $0.20\ \& \ 0.50\,\mathrm{m}$ & $787$ & $81.3$ & $81.8$ & $69.8$ \\
        $0.20\ \& \ 0.70\,\mathrm{m}$ & $2{,}451$ & $82.0$ & $89.4$ & $75.0$ \\
        $0.35\ \& \ 0.50\,\mathrm{m}$ & $723$ & $83.3$ & $83.0$ & $73.0$ \\
        $0.35\ \& \ 0.70\,\mathrm{m}$ & $2{,}383$ & $84.1$ & $89.5$ & $76.9$ \\
        $0.50\ \& \ 0.70\,\mathrm{m}$ & $1{,}683$ & $80.8$ & $88.7$ & $73.3$ \\
        \bottomrule
    \end{tabular}
\end{table}

\begin{table}[t]
    \centering
    \caption{Controlled hierarchical execution on R2R-CE using the same high-level VLM and reference-path planner.}
    \label{tab:hierarchical_navigation}
    \scriptsize
    \setlength{\tabcolsep}{2.5pt}
    \resizebox{\columnwidth}{!}{%
    \begin{tabular}{@{}lccccccc@{}}
        \toprule
        \textbf{Executor}
        & \textbf{SR}$\uparrow$
        & \textbf{OSR}$\uparrow$
        & \textbf{SPL}$\uparrow$
        & \textbf{NE}$\downarrow$
        & \textbf{CFR}$\uparrow$
        & \textbf{CFSR}$\uparrow$
        & \textbf{Avg.\ Cmds.}$\downarrow$\\
        \midrule
        StreamVLN-ZS & $90.4$ & $91.7$ & $53.9$ & $2.51$ & $46.4$ & $45.6$ & $5.18$\\
        DualVLN-ZS & $90.4$ & $91.4$ & $52.4$ & $2.46$ & $63.6$ & $62.6$ & $5.33$\\
        JanusVLN-ZS & $94.1$ & $94.3$ & $69.6$ & $2.18$ & $57.9$ & $57.7$ & $4.10$\\
        StreamVLN-FT & $95.0$ & $95.0$ & $74.4$ & $1.94$ & $62.2$ & $61.8$ & $4.32$\\
        PASS-VLN & $\mathbf{99.5}$ & $\mathbf{99.6}$ & $\mathbf{75.6}$ & $\mathbf{1.71}$ & $\mathbf{68.8}$ & $\mathbf{68.3}$ & $\mathbf{3.72}$\\
        \bottomrule
    \end{tabular}%
    }
\end{table}

\subsubsection{Body-Width-Conditioned Execution versus Passability Gating}

To investigate whether explicit binary passability information alone is sufficient for reliable execution, we compare against a decoupled GT Passability-Gated VLN baseline. This baseline is trained with a $0.10\,\mathrm{m}$ body width, making the training trajectories effectively all passable and removing width-dependent passability supervision. At test time, Habitat recomputes the NavMesh for the actual width and forces STOP when the remaining instructed path becomes impassable. As shown in Table~\ref{tab:width_generalization}, at larger body widths, the GT passability gate provides a larger SR advantage because more routes become impassable and can therefore be correctly terminated by the oracle gate. However, the width-agnostic VLN leads to more collisions and larger drops in CFR and CFSR. This shows that binary passability gating alone is insufficient: successful execution also requires body-width-dependent clearance, alignment, and avoidance. PASS-VLN incorporates these constraints through its body-conditioned geometry, allowing body width to affect both whether and how the route is executed.

\begin{table*}[t]
\centering
\caption{
Ablation of action representation, geometric input, and width conditioning.
Single-width variants are evaluated under the seen 0.35\,m embodiment,
while width-conditioned variants are additionally evaluated on seen and unseen body-width pairs.
}
\label{tab:ablation}
\setlength{\tabcolsep}{5.0pt}
\renewcommand{\arraystretch}{1.05}
\begin{tabular}{l c c c ccc cc}
\hline
\multirow{2}{*}{\textbf{Input / Method}}
& \multirow{2}{*}{\textbf{Action}}
& \multirow{2}{*}{\textbf{Train Widths}}
& \multirow{2}{*}{\textbf{Width Cond.}}
& \multicolumn{3}{c}{\textbf{Seen Width ($\mathbf{0.35}$\,m)}}
& \textbf{Seen PSR$\uparrow$}
& \textbf{Unseen PSR$\uparrow$} \\
\cline{5-7}
& & & &
\textbf{SR$\uparrow$} &
\textbf{CFR$\uparrow$} &
\textbf{CFSR$\uparrow$} &
$\mathbf{0.35\&0.70}$\,\textbf{m} &
$\mathbf{0.20\&0.50}$\,\textbf{m} \\
\hline

RGB (StreamVLN-FT)
& Atomic
& 0.35
& None
& 77.7 & 81.1 & 64.4
& -- & -- \\

RGB
& Latent
& 0.35
& None
& 79.4 & 88.1 & 71.4
& -- & -- \\

RGB + Depth
& Latent
& 0.35
& None
& 78.5 & 90.3 & 72.7
& -- & -- \\

RGB + Raw BEV
& Latent
& 0.35
& None
& 79.7 & 88.9 & 72.0
& -- & -- \\

RGB + Raw BEV + Width Prompt
& Latent
& 0.35 + 0.70
& Prompt
& 81.4 & 89.7 & 74.0
& 73.4 & 11.7 \\

RGB + Width-cond. BEV (PASS-VLN)
& Latent
& 0.35 + 0.70
& Geometry
& \textbf{83.4}
& \textbf{90.6}
& \textbf{76.1}
& \textbf{76.9}
& \textbf{69.8} \\

\hline
\end{tabular}
\end{table*}

\begin{figure*}[t]
    \centering
    \includegraphics[width=1.0\textwidth]{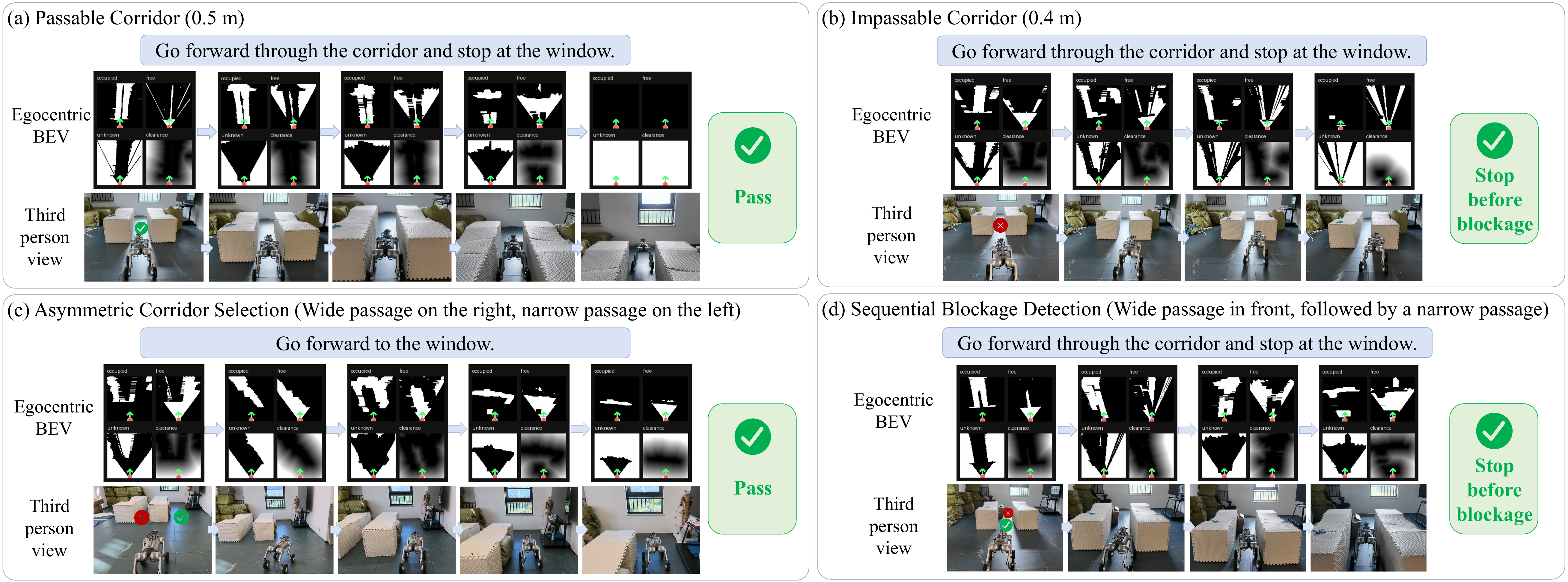}
    \caption{Real-world deployment on a Unitree Go2-W, demonstrating width-dependent traversal, blockage stopping, corridor selection, and sequential blockage detection.}
    \label{fig:real_world}
\end{figure*}

\begin{figure*}[t]
    \centering
    \includegraphics[width=1.0\textwidth]{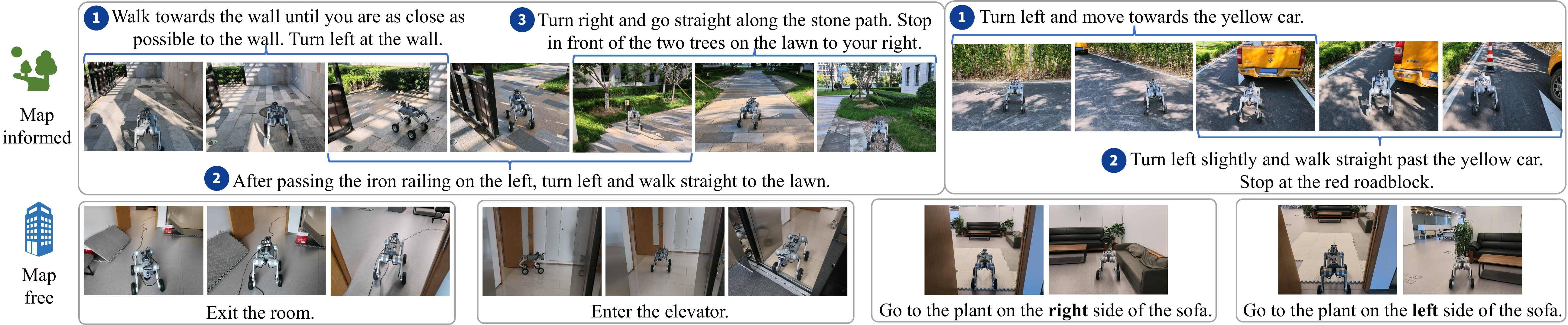}
    \caption{Real-world local-command execution in indoor and outdoor environments, where sequential map-informed commands support long-range navigation and map-free commands test local spatial grounding.}
    \label{fig:real_world_2}
\end{figure*}

\subsubsection{Paired Body-Width Evaluation}

Single-width performance alone cannot determine whether the model actually changes its decision according to body width. We therefore select paired episodes with identical scenes, instructions, start poses and reference routes, where the route is passable for a narrower body but impassable for a wider one. The narrow robot should therefore reach the instructed target, whereas the wide robot should stop before the first blockage. Table~\ref{tab:paired_width} confirms that PASS-VLN changes its behavior according to body width rather than following a width-independent strategy. 

\subsection{Local Execution in an Oracle Hierarchical Pipeline}

We further evaluate PASS-VLN as the local execution layer of a hierarchical navigation system on R2R-CE val\_unseen split. All methods share the same reference route and high-level pipeline. At each iteration, Habitat finds the next unreached waypoint on the reference path from the current robot pose and replans a shortest path to it. The local segment is then extended along the reference route, and a shared VLM (Gemma 4 31B~\cite{gemmateam2026gemma4}) generates a local command from the segment and egocentric visual evidence. After execution, the returned robot pose is used to plan the next segment and command. All methods use the same planner, VLM, and command-generation procedure, with at most $20$ local commands per episode.

As shown in Table~\ref{tab:hierarchical_navigation}, Avg.\ Cmds. is the average number of local commands per episode. PASS-VLN achieves the best overall performance, indicating that more reliable local execution reduces corrective interactions and improves the efficiency of hierarchical navigation. CFR and CFSR are lower than in single-command evaluation as they are computed over the long-horizon episode: a collision in any local command marks the whole episode as non-collision-free.

\subsection{Ablation Studies}
\label{subsec:ablations}

We perform controlled ablations on the map-informed subset to examine: (1) whether latent trajectories improve physical execution over atomic actions; (2) whether explicit geometric observations improve local physical execution; and (3) how body width should be incorporated into geometry for width-aware execution and generalization. Table~\ref{tab:ablation} compares six variants differing in input, action type, and training widths, where RGB + Raw BEV + Width Prompt explicitly provides the body width in the text prompt.

First, replacing atomic actions with latent trajectories improves CFR and CFSR, with a smaller gain in SR, showing that latent actions mainly improve collision-free execution. Second, depth and raw BEV provide explicit local geometry and improve physical execution to some extent, but remain independent of body width and therefore do not directly support width-dependent passability reasoning. Third, the form of width conditioning affects generalization to unseen body widths. With the same training widths, the RGB + Raw BEV + Width Prompt variant performs comparably on the seen pair but generalizes poorly to the unseen pair. This is because its narrow GSR drops to only $13.0\%$, indicating an overly conservative tendency to stop even when the route is passable. Under the same multi-width supervision, geometric conditioning therefore generalizes better than the evaluated width-prompt baseline.

\subsection{Real-World Deployment}
\label{subsec:real_robot}

Finally, we evaluate whether the local execution and body-width-aware passability capabilities transfer to a physical robot. We deploy PASS-VLN on a Unitree Go2-W ($0.70\times0.43\times0.50\,\mathrm{m}$), using a configured width of $0.45\,\mathrm{m}$ including a $0.01\,\mathrm{m}$ safety margin. RGB and depth are provided by the onboard camera and a RealSense D435i, respectively, with inference running on an external server.

As shown in Fig.~\ref{fig:real_world}, we first evaluate body-width-aware passability by varying the width of a mandatory passage from $1.0$ to $0.4\,\mathrm{m}$ in $0.1\,\mathrm{m}$ increments, with three trials per setting. PASS-VLN traverses $0.6$--$1.0\,\mathrm{m}$ passages collision-free in all trials, while at $0.5\,\mathrm{m}$ it traverses in two of three trials but with minor boundary contact. At $0.4\,\mathrm{m}$, it consistently stops before the blockage. It also selects the feasible alternative in asymmetric layouts and detects downstream blockages in sequential passages, demonstrating consistent width-dependent passability reasoning on the real robot. This design may also support changes in the effective lateral footprint, such as added payloads, through an updated width configuration.

We further evaluate local-command execution in an approximately $500\,\mathrm{m}^2$ indoor workspace and a $3600\,\mathrm{m}^2$ outdoor environment as shown in Fig.~\ref{fig:real_world_2}. For map-informed navigation, sequential local commands guide the robot over the large area, demonstrating that the local commands can be composed for extended navigation. For map-free commands, PASS-VLN successfully grounds nearby landmarks and spatial relations, demonstrating local instruction understanding and execution in the tested environments.

\section{Conclusion}
\label{sec:conclusion}

We presented PASS-VLN, a passability-aware local-command VLN benchmark and local navigation model for body-width-dependent physical execution. By combining body-width-conditioned geometric reasoning with latent-action prediction, PASS-VLN enables an agent to reach feasible targets while stopping before body-width-induced blockages, even at previously unseen body widths. Experiments further demonstrate its effectiveness as a local executor in reference-route-assisted hierarchical navigation and real-world deployment. Future work will further improve local-command success and collision robustness through finer-grained stopping supervision, collision-free action learning, and more robust geometry-aware control, particularly for wide embodiments and narrow-clearance scenarios.

\bibliographystyle{IEEEtran}
\bibliography{references}

\end{document}